\documentclass[11pt,a4paper]{article}
\usepackage[hyperref]{acl2020}
\usepackage{times}
\usepackage{latexsym}
\usepackage{multirow}
\usepackage{graphicx}
\usepackage{amsmath}
\usepackage{caption}
\usepackage{float}

\usepackage{microtype}

\aclfinalcopy % Uncomment this line for the final submission
\def\aclpaperid{9} %  Enter the acl Paper ID here

\title{Using Large Pretrained Language Models for Answering User Queries from Product Specifications}

\author{
Kalyani Roy$^1$,  Smit Shah$^1$\thanks{\hspace{2mm}Work done while author was at IIT Kharagpur.}
, Nithish Pai$^2$, Jaidam Ramtej$^2$, Prajit Prashant Nadkarn$^2$, \\
\bf {Jyotirmoy Banerjee$^2$, 
Pawan Goyal$^1$, \and Surender Kumar$^2$} \\
$^1$Indian Institute of Technology Kharagpur\\
$^2$Flipkart\\
kroy@iitkgp.ac.in, smitsunny11@gmail.com, 
\{nithish.p, jaidam.ramtej, prajit.pn \}@\\flipkart.com, jyoban@gmail.com, pawang@cse.iitkgp.ac.in,
surender.k@flipkart.com
}

\begin{document}
\maketitle
\begin{abstract}
While buying a product from the e-commerce websites, customers generally have a plethora of questions. From the perspective of both the e-commerce service provider as well as the customers, there must be an effective question answering system to provide immediate answers to the user queries. While certain questions can only be answered after using the product, there are many questions which can be answered from the product specification itself. Our work takes a first step in this direction by finding out the relevant product specifications, that can help answering the user questions. We propose an approach to automatically create a training dataset for this problem. We utilize recently proposed XLNet and BERT architectures for this problem and find that they provide much better performance than the Siamese model, previously applied for this problem~\cite{lai2018simple}. Our model gives a good performance even when trained on one vertical and tested across different verticals. 

\end{abstract}
\section{Introduction}

\begin{figure}[t]
	\centering
	\includegraphics[width=0.48\textwidth]{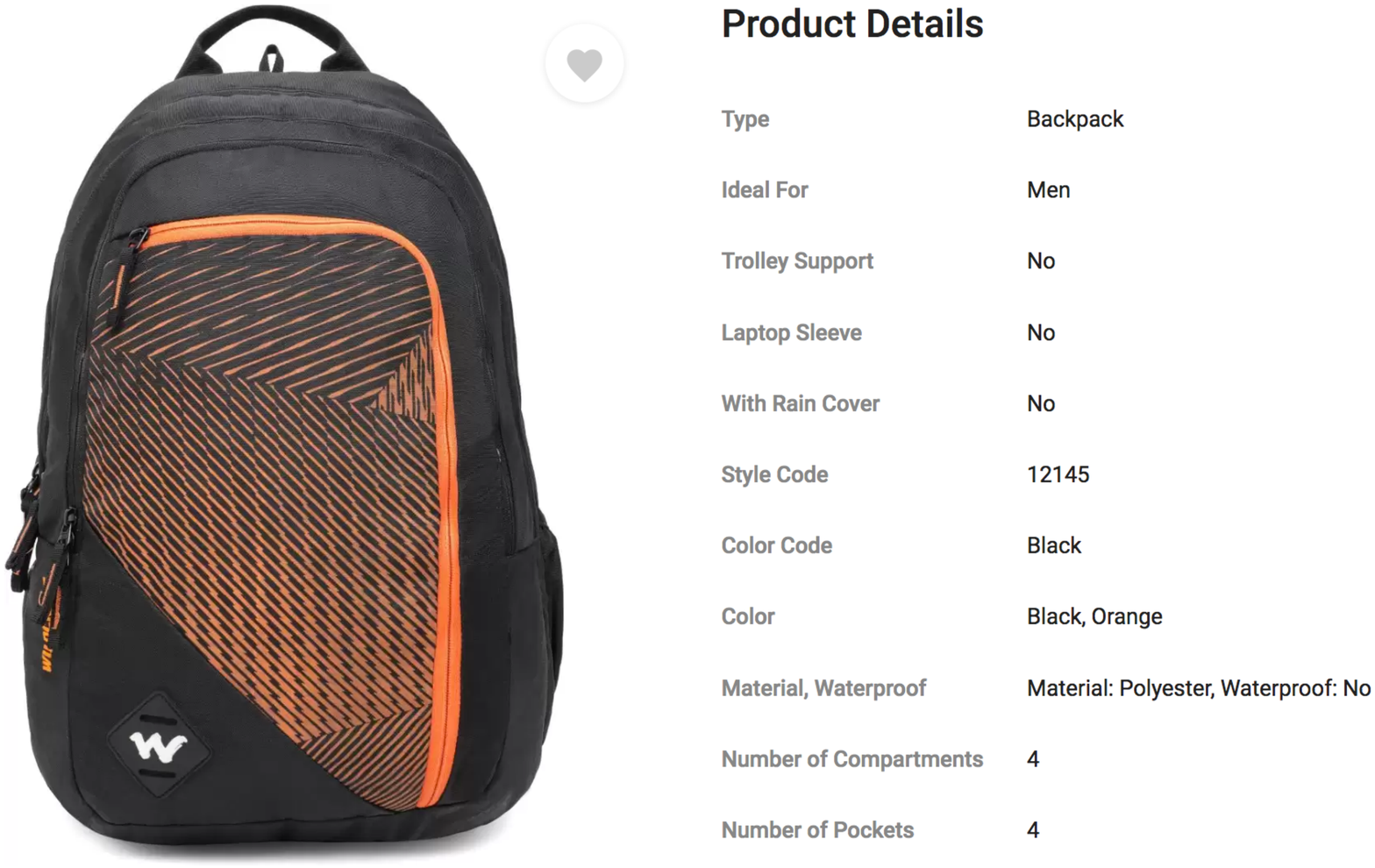}
	\caption{Snapshot of a product with its specifications.}
	\label{fig:q-spec}
\end{figure}

Product specifications are the attributes of a product. These specifications help a user to easily identify and differentiate products and choose the one that matches certain specifications. There are more than $80$ million products across $80+$ product categories on Flipkart~\footnote{Flipkart Pvt Ltd. is an e-commerce company based in Bangalore, India.}.
The $6$ largest categories are - \textit{Mobile}, \textit{AC}, \textit{Backpack}, \textit{Computer}, \textit{Shoes}, and \textit{Watches}. A large fraction of user queries ($\sim 20\%$)\footnote{We randomly sampled $1500$ questions from all these verticals except \textit{Mobile} and manually annotated them as to whether these can be answered through product specifications.} can be answered with the specifications. Product specifications would be helpful in providing instant responses to questions newly posed by users about the corresponding product. Consider a question ``What is the fabric of this bag?" This new question can be easily answered by retrieving the specification ``Material" as the response. Fig.~\ref{fig:q-spec} depicts this scenario.

Most of the recent works on product related queries on e-commerce leverage the product reviews to answer the questions%by posing it as a Question and Answer (QA) like task
~\cite{gao2019product,zhao2019riker,mcauley2016addressing}. Although reviews are a rich source of data, they are also subject to personal experiences. People tend to give many reviews on some products and since it is based upon their personal experience, the opinion is also diverse. This creates a massive volume and range of opinions and thus makes review systems difficult to navigate. Sometimes products do not even have any reviews that can be used to find an answer, also the reviews do not mention the specifications a lot, but mainly deal with the experience. So, there are several reasons why product specifications might be a useful source of information to answer product-related queries which does not involve user experience to find an answer. As the specifications are readily available, users can get the response instantly.

This paper attempts to retrieve the product specifications that would answer the user queries. While solving this problem, our key contributions are as follows - (i) We demonstrate the success of XLNet on finding product specifications that can help answering product related queries. It beats the baseline Siamese method by $0.14-0.31$ points in HIT@1. (ii) We utilize a method to automatically create a large training dataset using a semi-supervised approach, that was used to fine-tune XLNet and other models. 
(iii) While we trained on \textit{Mobile} vertical, we tested on different verticals, namely, $\mathit{AC}$, $\mathit{Backpack}$, $\mathit{Computer}$, $\mathit{Shoes}$,  $\mathit{Watches}$, which show promising results.
% HIT : human intelligence tasks 

\begin{table}[t]
	\begin{center}
		\small
		\setlength{\tabcolsep}{7pt}
		%\begin{adjustbox}{max width=\textwidth}
		\begin{tabular}{l|c|r|c} % <-- Alignments: 1st column left, 2nd middle and 3rd right, with vertical lines in between
			\hline
			\textbf{Dataset} & \textbf{Products} & \textbf{Questions} & \textbf{Avg. Specs}\\
			\hline
			\textit{Mobile} & 1,175 & 260,529 & 55\\
			$\mathit{AC}$ & 300 & 16,545 & 35\\
			$\mathit{Backpack}$ & 300  & 16,878 & 17\\
			$\mathit{Computer}$ & 300  & 93,589 & 60\\
			$\mathit{Shoes}$ & 300  & 5,812 & 10\\
			$\mathit{Watches}$ & 300  & 21,392 & 50\\
			\hline
		\end{tabular}
		
		\caption{Statistics of 6 largest categories.}% used in this paper.}
		\label{tab:data}
	\end{center}
	
\end{table}

\section{Background and Related Work}

In recent years, e-commerce product question answering (PQA) has received a lot of attention. %Most of the existing strategies aim at extracting relevant sentences from reviews to answer the given question.  
~\citet{yu2018responding} present a framework to answer product related questions by retrieving a ranked list of reviews and they employ the Positional Language Model (PLM) to create the training data. ~\citet{chen2019answer} apply a multi-task attentive model to identify plausible answers.
~\citet{lai2018simple} propose a Siamese deep learning model for answering questions regarding product specifications. The model returns a score for a question and specification pair. %For creating the dataset, they used Amazon Mechanical Turk.
~\citet{mcauley2016addressing} exploit product reviews for answer prediction via a Mixture of Expert (MoE) model. This MoE model makes use of a review relevance function and an answer prediction function. %One restriction of this model is that it can only be used for answer selection given a candidate answer set. 
%Although the question answer collections and review collections are involved in the learning procedures, one assumption is that a 
It assumes that a candidate answer set containing the correct answers is available for answer selection.% Such setting is not practical for instant response generation.
~\citet{cui2017superagent} develop a chatbot for e-commerce sites known as SuperAgent. SuperAgent considers question answer collections, reviews and specifications when answering questions. It selects the best answer from multiple data sources. Language representation models like BERT~\cite{devlin2019bert} and XLNet~\cite{yang2019xlnet} are pre-trained on vast amounts of text
and then fine-tuned on task-specific labelled data. The resulting models have achieved state of the art in many natural language processing tasks including question answering. ~\citet{dzendzik2019dish} employ BERT to answer binary questions by utilizing customer reviews. 

In this paper, unlike some of the previous works~\cite{lai2018simple, chen2019answer} on PQA that solely rely on human annotators to annotate the training instances, we propose a semi-supervised method to label training data. We leverage the product specifications to answer user queries by using BERT and XLNet. 

 \iffalse
 \begin{table}[t]
 	%	\setlength{\tabcolsep}{4pt}
 	\centering
 	\begin{tabular}{c|c|c}
 		\hline
 		\textbf{Upper} & \textbf{Lower} & \textbf{Accuracy on } \\
 		\textbf{Limit $\theta$} & \textbf{Limit $\beta$} & \textbf{validation data}  \\ \hline
 		0.52 & 0.28  & 0.64 \\
 		0.52 & 0.34  & 0.67 \\
 		\textbf{0.34} & \textbf{0.34} & \textbf{0.72} \\
 		\hline
 	\end{tabular}
 	\caption{Threshold values to label training dataset.} %Finding the best t
 	\label{tab:threshold}
 \end{table}
 
\fi

\section{Problem Statement}
Here, we formalize the problem of answering user queries from product specifications. Given a question $Q$ about a product $P$ and the list of $M$ specifications $\{s_1$, $s_2$, ..., $s_M\}$ of $P$, our objective is to identify the specification $s_i$ that can help answer $Q$. Here, we assume that the question is answerable from specifications.%, where $M$ is the number of specifications of $P$, and $s_i$ is the $i^{th}$ specification of $P$. 

\section{Model Architecture}
%To make our objectives more concrete, 

Our goal is to train a classifier that takes a question and a specification as input (e.g., ``Color Code Black'') and predicts whether the specification is relevant to the question. 
% whether the specification is relevant to the question. % During evaluation, for a given question of a product, all of the candidate specifications will receive a score from the classifier based upon its relevancy to the question. %The top-ranked specification will be considered as the answer to the question. 
We take Siamese architecture~\cite{lai2018simple} as our baseline method. We fine-tune BERT and XLNet for this classification task.

\textbf{Siamese:} We train a 100-dimensional word2vec embedding on the whole corpus (all questions and specifications as shown in Table \ref{tab:data}.) to get the input word representation. In the Siamese model, the question and specification is passed through a Siamese Bi-LSTM layer. Then we use max-pooling on the contextual representations to get the feature vectors of the question and specification. We concatenate the absolute difference and hadamard product of these two feature vectors and feed it to two fully connected layers of dimension 50 and 25, subsequently. Finally, the softmax layer gives the relevance score. % We trained a 100 dimensional word embedding on our whole dataset.
%\PG{Need more description. What was the input embedding? How was that obtained? Was this model exactly same as Lai et al.?}

%\vspace{-2mm}
\textbf{BERT} and \textbf{XLNet :} %An input sequence for BERT consists of two ``sentences'', each ends with a [SEP] token. And a [CLS] token is added to the head, whose corresponding hidden states are used to make the next sentence prediction. 
The architecture we use for fine-tuning BERT and XLNet is the same. We begin with the pre-trained BERT\textsubscript{Base} and XLNet\textsubscript{Base} model. % which represents the normal model variant. 
To adapt the models for our task, we introduce a fully-connected layer over the final hidden state corresponding to the [CLS] input token. During fine-tuning, we optimize the entire model end-to-end, with the additional softmax classifier parameters $W\in R^{K \times H}$, where $H$ is the dimension of the hidden state vectors and $K$ is the number of classes.

\begin{table}[!t]
	\setlength{\tabcolsep}{7pt}
	\small
	\centering
	
	\begin{tabular}{l|c|c|c|c}
		\hline
		\textbf{Dataset} & \textbf{ \# que-spec}  & \multicolumn{3}{c}{\textbf{Answer type (in \%)}} \\ \cline{3-5}
		& \textbf{pairs}
		& \textbf{Num} & \textbf{Y/N} & \textbf{Other}\\\hline
		\textit{AC}& 3693 &  0.27 & 0.52 & 0.21\\
		\textit{Backpack}& 2693 & 0.29 & 0.48 & 0.23 \\
		\textit{Computer}& 2718 & 0.04 & 0.78 & 0.18 \\
		\textit{Shoes}& 999  & 0.09 & 0.49 & 0.42 \\
		\textit{Watches}& 1700 & 0.17 & 0.59 & 0.24\\ \hline
	\end{tabular}
	\caption{Test datasets statistics.}% Neumerica}
	\label{tab:test_data_stat}
\end{table}

\section{Experimental Setup}

\subsection{Dataset Creation} \label{sec:data_creation}
The Statistics for the 6 largest categories used in this paper are shown in Table \ref{tab:data}, containing a snapshot of product details up to January 2019. Except for mobiles, for other domains, 300 products were sampled.
As the number of question-specification pairs is huge, manually labelling a sufficiently large dataset is a tedious task. So, we propose a semi-supervised method to create a large training dataset using Dual Embedding Space model (DESM)~\cite{mitra2016dual}.
%DESM leverages question output embedding spaces and specification output embedding spaces
%to derive richer distributional relationships. %During ranking, they map the query words into the input space and the document words into the output space, and compute a query-document relevance score by aggregating the cosine similarities across all the query-document word pairs. Mitra  et al.

Suppose a product $P$ has $S$ specifications and $Q$ questions. For a question $q_{i} \in Q$ and a specification $s_{j} \in S$, we find dual embedding score $DUAL(q_{i},s_{j})$ using Equation~\ref{equ:dual}, where $t_q$ and $t_s$ denote the vectors for the question and specification terms, respectively. We consider $(q_{i},s_{j})$ pair positive if $DUAL(q_{i},s_{j}) \geq\theta$ %\geq
%. We consider $(q_{i},s_{j})$ pair 
and negative if $DUAL(q_{i},s_{j}) < \theta$. 
%So, for the product P we have $S \times Q$ question-specification pairs and exactly Q of them are positive pairs. 
\begin{equation}
    DUAL (q_i, s_j) = \frac{1}{ |q_i| }\sum_{t_q \in q_i}\frac{{t_q}^T \overline{s_j}}{\parallel t_q \parallel \parallel \overline{s_j}\parallel} \label{equ:dual}
\end{equation}
where
\begin{equation}
 \overline{s_j} = \frac{1}{|s_j|} \sum_{t_s \in s_j} \frac{t_s}{\parallel t_s \parallel}
\end{equation}

\begin{table}[t]
\small
	\centering	
	\setlength{\tabcolsep}{4pt}
	\begin{tabular}{l|c|ccc} \hline
		Dataset & Model & HIT@1 & HIT@2 & HIT@3  \\ \hline
		
		\multirow{5}{*}{ \textit{AC}} 
		& BERT-380 & 0.05 & 0.09 & 0.14 \\
		& XLNet-380 & 0.20 &0.32 & 0.39\\ \cline{2-5}
		& Siamese & 0.38 & 0.53 &0.61  \\
		& BERT & 0.62 & \textbf{0.77} & \textbf{0.81}\\
		& XLNet & \bfseries{0.69} & \bfseries{0.77} & 0.80  \\ \hline
		
		\multirow{5}{*}{ \textit{Backpack}} 
		& BERT-380 & 0.17 & 0.27 & 0.34 \\
		& XLNet-380 & 0.27 & 0.41 & 0.48 \\ \cline{2-5}
		& Siamese & 0.35 & 0.53 & 0.65 \\
		& BERT &\textbf{ 0.50} & 0.66 & 0.69  \\
		& XLNet & 0.49 & \textbf{0.67} & \textbf{0.70}  \\ \hline
		
		\multirow{5}{*}{ \textit{Computer}} 
		& BERT-380 & 0.14 & 0.16 & 0.22 \\
		& XLNet-380  & 0.06 & 0.16 & 0.18 \\ \cline{2-5}
		& Siamese & 0.5 &0.6 & 0.72 \\
		& BERT & 0.68 & 0.80 & 0.90  \\
		& XLNet & \textbf{0.70} & \textbf{0.86} & \textbf{0.92}  \\ \hline
		
		\multirow{5}{*}{ \textit{Shoes}} 
		& BERT-380 &  0.22 & 0.40 & 0.55\\
		& XLNet-380 & 0.25 & 0.45 & 0.60  \\ \cline{2-5}
		& Siamese & 0.42 & 0.55 & 0.62  \\       
		& BERT & 0.60 & 0.72 & 0.84 \\
		& XLNet & \textbf{0.63} & \textbf{0.77} & \textbf{0.88}  \\ \hline
		
		\multirow{5}{*}{ \textit{Watches}} 
		& BERT-380  & 0.05 & 0.09 & 0.15\\
		& XLNet-380 & 0.24 &  0.36 & 0.45  \\ \cline{2-5}
		& Siamese  & 0.42 & 0.65 & 0.69 \\
		& BERT & 0.54 & 0.60 & 0.74  \\
		& XLNet & \textbf{0.60} & \textbf{0.76} & \textbf{0.84} \\ \hline
	\end{tabular}
	\caption{Performance comparison of different models.}
	\label{tab:test_result}
	\vspace{-1.5mm}
\end{table}

We take $Mobile$ dataset to create labelled training data since most of the questions come from this vertical. We choose the threshold value ($\theta$) which gives the best accuracy on manually labelled balanced validation dataset consisting of $380$ question and specification pairs. %The validation dataset is balanced i.e., it has same number of positive and negative data.
We train a word2vec%~\cite{mikolov2013efficient}
~\cite{mikolov2013nips} model on our training dataset to get the embeddings of the words. The word2vec model learns two weight matrices during training. The matrix corresponding to the input space and the output space is denoted as IN and OUT word embedding space respectively. Word2vec leverages only the input embeddings (IN), but discards the output embeddings (OUT), whereas DESM utilizes both IN and OUT embeddings.
To compute the DUAL score of a question and specification, we take OUT-OUT vectors as it gives the best validation accuracy. We find that for $\theta=0.34$, we gain maximum accuracy value of $0.72$ on the validation set. This creates a labelled training dataset $\mathcal{D}$ with $57,138$ positive pairs and $655,290$ negative pairs. For training, we take all the positive data from $\mathcal{D}$ and we randomly sample an equal number of negative examples from $\mathcal{D}$.

\iffalse. Table~\ref{tab:threshold} shows the accuracy on the validation data for different values of $\theta$ and $\beta$.\fi

\begin{table*}[t]
	\centering
	\small
	\begin{tabular}{|p{.13\textwidth}|p{.25\textwidth}|p{.25\textwidth}|p{.25\textwidth}|}
		\hline
		\textbf{Question} & \textbf{Siamese} & \textbf{BERT} & \textbf{XLNet} \\ \hline \hline
		\multirow{3}{*}{\parbox{.9\linewidth}{Is it single core or multi core?}}
		& processor name core i3  & internal mic single digital microphone &\textbf{number of cores 2}\\ \cline{2-4}
		& processor variant 7100u & processor name core i3 & processor name core i3\\ \cline{2-4}
		& os architecture 64 bit &\textbf{number of cores 2} &processor brand intel \\  \hline
		\multirow{3}{*}{\parbox{.9\linewidth}{Does 16 inch laptop fit in to it?}}
		& depth 13 inch & \textbf{compatible laptop size 15.4 inch} & \textbf{compatible laptop size 15.4 inch} \\ \cline{2-4}
		& width 9 inch & laptop sleeve no & depth 13 inch \\ \cline{2-4}
		& height 19 inch & depth 13 inch & height 19 inch \\ \cline{2-4}  \hline
		%\multirow{3}{*}{\parbox{.9\linewidth}{Is this 5 star or 3 star?}}
		%& \textbf{star rating 3 star bee rating}  & \textbf{star rating 3 star bee rating} &\textbf{star rating 3 star bee rating}\\ \cline{2-4}
		%& refrigerant r-410a & outdoor w$\times$h$\times$d 27 cm $\times$ 54.5 cm $\times$ 28.5 cm & outdoor w$\times$h$\times$d 27 cm $\times$ 54.5 cm $\times$ 28.5 cm\\ \cline{2-4}
		%& condenser coil copper & 1 year warranty on product & indoor w$\times$h$\times$d 10 cm $\times$ 19.8 cm $\times$ 30 cm\\  \hline
	\end{tabular}
	%\captionsetup{justification=centering}
	\caption{Top three specifications returned by different models for two questions. Correct specification is highlighted in bold.}
	\label{tab:result_example}
\end{table*}

To create the test datasets, domain experts manually annotate the correct specification for a question. 
%We keep all dataset except $Mobile$ for testing. %We divide the datasets in such a way that 
As the test datasets come from different verticals, there is no product in common with the training set. The details of different test datasets are shown in Table~\ref{tab:test_data_stat}. 
We analyze the questions in the test datasets and find that the questions can be roughly categorized into three classes - numerical, yes/no and others based upon the answer type of the questions. For a question, we have a number of specifications and only one of them is correct.

\subsection{Training and Evaluation}
We split the \textit{Mobile} dataset into 80\% and 20\% as training set and development set, respectively. The Siamese model is trained for 20 epoch with Stochastic Gradient Descent optimizer and learning rate 0.01. The fine-tuning of BERT and XLNet is done with the same experimental settings as given in the original papers. In all the models, we minimize the cross-entropy loss while training. BERT-380 and XLNet-380 models are fine-tuned on the 380 labeled validation dataset that was used for creating the training dataset in Section ~\ref{sec:data_creation}.

%\subsection{Evaluation Metrics}
During evaluation, we sort the question specification pairs according to their relevance score. From this ranked list, we compute whether the correct specification appears within top $k, k \in\{1,2,3\}$ positions. The ratio of correctly identified specifications in top $1$, $2$, and $3$ positions to the total number of questions is denoted as HIT@1, HIT@2 and HIT@3 respectively. 

%Then we calculate for each questions how many specifications appear in % Then we compute how many question specification pairs are correctly identified in 
%top $1, 2, 3$ positions and we call this HIT@1, HIT@2 and HIT@3 respectively. %We also determine the mean reciprocal rank (MRR).

\section{Results and Discussion}

Table~\ref{tab:test_result} shows the performance of the models on different datasets\footnote{Unsupervised DUAL embedding model gave very similar results to Siamese model, and is not reported.}. BERT-380 and XLNet-380 perform very poorly, but when we use the train dataset created with DESM, there is a large boost in the models' performance and it shows the effectiveness of our semi-supervised method in generating labeled dataset. Both BERT and XLNet outperform the baseline Siamese model~\cite{lai2018simple} by a large margin, and retrieve the correct specification within top 3 results for most of the queries. For \textit{Backpack} and \textit{AC}, both BERT and XLNet are very competitive. XLNet outperforms BERT in \textit{Computer}, \textit{Shoes}, and \textit{Watches}. Only in HIT@1 of \textit{AC}, BERT has surpassed XLNet with $0.07$ points. We see that all the models have performed better in \textit{Computer} compared to the other datasets. \textit{Computer} has the highest percentage of yes/no questions and  this might be one of the reasons, as some questions might have word overlap with correct specification.  Table~\ref{tab:result_example} shows the top three specifications returned by different models for some questions. We see that Siamese architecture returns results which look similar to na\"{i}ve word match, and retrieve wrong specifications. On the other hand, BERT and XLNet are able to retrieve the correct specifications.
%if the question does not have any word overlap with the specification, BERT struggles to retrieve the correct specification at the top.

\textbf{Error Analysis:} We assume that for each question, there is only one correct specification, but %some queries contain multiple questions together, e.g., ``is the keyboard backlight? And how is the battery backup?". In that case, only one question is answered, even though the answer of all the questions might be found from specifications. In some questions, 
the correct answer may span multiple specifications and our models can not provide a full answer. For example, in \textit{Backpack} dataset, the dimension of the backpack, i.e., its height, weight, depth is defined separately. So, when user queries about the dimension, only one specification is provided. Some specifications are given in one unit, but users want the answer in another unit, e.g., ``what is the width of this bag in cms?". Since the specification is given in inches, the models show the answer in inches. So, the answer is related, but not exactly correct. Users sometimes want to know the difference between certain specification types, what is meant by some specifications. For example, consider the questions ``what is the difference between inverter and non-inverter AC?", ``what is meant by water resistant depth?". While we can find the type of inverter, the water resistant depth of a watch etc. from specifications, the definition of the specification is not given. As we have generated train data labels in semi-supervised fashion, it also contributes to inaccurate classification in some cases.

\section{Conclusion and Future Work}

In this paper, we proposed a method to label training data with little supervision. We demonstrated that large pretrained language models such as BERT and XLNet can be fine-tuned successfully to obtain product specifications that can help answer user queries. We also achieve reasonably good results even while testing on different verticals.

We would like to extend our method to take into account multiple specifications as an answer. We also plan to develop a classifier to identify which questions can not be answered from the specifications. %As an extension of our work, we plan to use the questions and answers provided by users, along with the specifications, for the given task. 
%Specifically, we proposed a method to label training data with little supervision, and showed that even while testing on different verticals, we achieve reasonably good results.

%One direction of future work involves having the confidence on the answer. 

%\input{template.tex}
%\bibliography{anthology,acl2020}
\bibliography{acl2020}
\bibliographystyle{acl_natbib}

\end{document}